\documentclass[conference, a4paper]{IEEEtran}
\IEEEoverridecommandlockouts
\usepackage{cite}
\usepackage{amsmath,amssymb,amsfonts}
\usepackage{algorithmic}
\usepackage{graphicx}
\usepackage{textcomp}
\usepackage{xcolor}
\usepackage[justification=centering]{caption}
\usepackage{subcaption}
\usepackage{blindtext}
\usepackage{placeins}
\usepackage[colorlinks=true,linkcolor=red]{hyperref}%
\usepackage{etoolbox}
\usepackage{multicol}
\usepackage{multirow}
\makeatletter
\patchcmd{\@makecaption}
  {\scshape}
  {}
  {}
  {}
\makeatother
\def\BibTeX{{\rm B\kern-.05em{\sc i\kern-.025em b}\kern-.08em
    T\kern-.1667em\lower.7ex\hbox{E}\kern-.125emX}}

\begin{document}

\title{LAPFormer: A Light and Accurate Polyp Segmentation Transformer\\

}

\author{
\IEEEauthorblockN{Mai Nguyen, Tung Thanh Bui, Quan Van Nguyen, Thanh Tung Nguyen, Toan Van Pham}
\IEEEauthorblockA{R\&D Lab, Sun* Inc\\
\{nguyen.mai, bui.thanh.tung, nguyen.van.quan, nguyen.tung.thanh, pham.van.toan\}@sun-asterisk.com\\}
}

\maketitle

\begin{abstract}
Polyp segmentation is still known as a difficult problem due to the large variety of polyp shapes, scanning and labeling modalities. This prevents deep learning model to generalize well on unseen data. However, Transformer-based approach recently has achieved some remarkable results on performance with the ability of extracting global context better than CNN-based architecture and yet lead to better generalization. To leverage this strength of Transformer, we propose a new model with encoder-decoder architecture named LAPFormer, which uses a hierarchical Transformer encoder to better extract global feature and combine with our novel CNN (Convolutional Neural Network) decoder for capturing local appearance of the polyps. Our proposed decoder contains a progressive feature fusion module designed for fusing feature from upper scales and lower scales and enable multi-scale features to be more correlative. Besides, we also use feature refinement module and feature selection module for processing feature. We test our model on five popular benchmark datasets for polyp segmentation, including Kvasir, CVC-Clinic DB, CVC-ColonDB, CVC-T, and ETIS-Larib.
\end{abstract}


\section{Introduction}
\subsection{Overview}
Colorectal cancer (CRC) is the most common cancer around the world \cite{bernal2017comparative}.
Colonoscopy has always been recognised as the standard diagnostic for the early detection of colorectal cancer. Therefore, several deep learning methods have been proposed to aid clinical system in identifying colonic polyps. Among them, segmentation approach is significantly considered as the most appropriate way with promising result recently. However, colonoscopy has some limitations. In some previous reports, about 18\% of polyps are missed from the diagnosis process \cite{kim2017miss, lee2017risk}. This is because it is an operator-driven procedure and solely dependent on the knowledge and skills of the endoscopist. With the current colonoscopy equipment, less experienced endoscopists cannot distinguish polyp regions during colonoscopy examinations \cite{wadhwa2020physician}. More importantly, previous research has shown that increasing polyp detection accuracy by 1\% reduces colorectal cancer risk by approximately 3\%. Therefore, improving polyp detectability and robust segmentation tools are important in this problem.

Recently, with the vigorous development of deep learning technology, the accuracy of many classical problems has improved, including image segmentation problems. Various studies aimed to develop CADx models for automatic polyp segmentation. There are also a few studies aiming to build a specific model for polyp segmentation. HarDNet-MSEG \cite{huang2021hardnet} is one of them, which is an encoder-decoder architecture base on HarDNet \cite{chao2019hardnet} backbone that achieved high performance on the Kvasir-SEG dataset with processing speed up to 86 FPS. AG-ResUNet++ improved UNet++ with attention gates and ResNet backbone. Another study called Transfuse combined Transformer and CNN using BiFusion module \cite{zhang2021transfuse}. ColonFormer \cite{duc2022colonformer} used MiT backbone, Uper decoder, and residual axial reverse attention to further boost the polyp segmentation accuracy. NeoUNet \cite{ngoc2021neounet} and BlazeNeo \cite{an2022blazeneo} proposed effective encoder-decoder networks for polyp segmentation and neoplasm detection. Generally, research in changing model architecture is still a potential approach.

Among the recent deep learning architecture, Transformer based architecture has attracted the most attention. To efficient semantic segmentation, incorporating the advantages of a hierarchical Transformer encoder with a suitable decoder head has been widely researched. In this paper, we utilize a Transformer backbone as an encoder and propose a novel, light, and accurate decoder head for polyp segmentation task. The proposed decode head consists of a light feature fusion module which efficiently reduces the semantic gaps between feature from two level scales, and a feature selection module along with a feature refinement module which help handling feature from backbone better and calibrating feature comes from progressive feature fusion module before prediction.

\subsection{Our contributions}
Our main contributions are:
\begin{itemize}
    \item We propose a \textbf{L}ight and \textbf{A}ccurate \textbf{P}olyp Sementation \textbf{Transformer}, called \textbf{LAPFormer}, that integrates a hierarchical Transformer backbone as encoder.
    \item A novel decoder for LAPFormer, which leverages multi-scale features  and consists of Feature Refinement Module, Feature Selection Module to produce fine polyp segmentation mask.
    \item Extensive experiments indicate that our LAPFormer achieves state-of-the-art on CVC-ColonDB \cite{tajbakhsh2015automated} and get competitive results on different famous polyp segmentation benchmarks while being less computation complexity than other Transformer-based methods.
\end{itemize}


\section{Related Works}

\subsection{Semantic Segmentation}
Semantic segmentation is one of the essential tasks in computer vision which is required to classify each pixel in the image. Recently, deep learning had an enormous impact on computer vision field, including semantic segmentation task. Many deep learning models are based on fully convolutional networks (FCNs) \cite{long2015fully}, which encoder gradually reduces the spatial resolution and captures more semantic context of an image with larger receptive fields. However, CNN still has small receptive fields, leading to missing context features. To overcome this limitation, PSPNet \cite{zhao2017pyramid} proposed Pyramid Pooling Module to enhance global context, DeepLab \cite{chen2017deeplab} utilized atrous convolution to expand receptive fields.

\subsection{Vision Transformer}
Transformer \cite{vaswani2017attention} is a deep neural network architecture, originally proposed to solve machine translation task in natural language processing. Nowadays, Transformer has a huge influence on natural language processing and other tasks, including computer vision. Vision Transformer (ViT) \cite{dosovitskiy2020image} is the first model successfully apply Transformer in computer vision by dividing an image into sequences and treating each sequence as a token, and then putting them into Transformer. Following this success, PVT \cite{wang2021pyramid}, Swin \cite{liu2021swin}, and SegFormer \cite{xie2021segformer} are designed as hierarchical Transformer, generating feature maps at different scales to enhance the local feature, thus improving the performance in dense prediction tasks, including detection and segmentation.

\subsection{Polyp segmentation}
Recent advances in deep learning have helped solve medical semantic segmentation tasks effectively, including polyp segmentation. However, it remains a challenging problem due to medical image characteristics, and polyps come in different sizes, shapes, textures, and colors. U-net \cite{ronneberger2015u} is a well-known medical image segmentation model consisting of residual connections to preserve local features between encoder and decoder. Inspired by U-net, most models \cite{srivastava2021gmsrf}, \cite{zhou2018unet++}, \cite{jha2019resunet++}, \cite{huang2021hardnet}, \cite{wei2021shallow}, \cite{zhao2021automatic}, for medical image segmentation use an architecture containing a CNN backbone as an encoder, and a decoder tries to fuse features at different scales to generate segmentation result. PraNet \cite{fan2020pranet}, SFA \cite{fang2019selective} focus on the distinction between a polyp boundary and background to improve segmentation performance. CaraNet \cite{lou2021caranet} designs a attention module for small medical object segmentation. Since the successful application of Transformer in computer vision, current methods \cite{wang2022stepwise}, \cite{dong2021polyp}, \cite{duc2022colonformer} have utilized the capability of Transformer as a backbone for polyp segmentation task and yielded promising results. Other methods \cite{chen2021transunet}, \cite{zhang2021transfuse}, \cite{sanderson2022fcn} use both Transformer and CNN backbone for feature extraction and combine features from two brands, thus enhancing segmentation result.

\section{Proposed Method}

\begin{figure*}[t]
    \centering
    \includegraphics[width=0.9\linewidth]{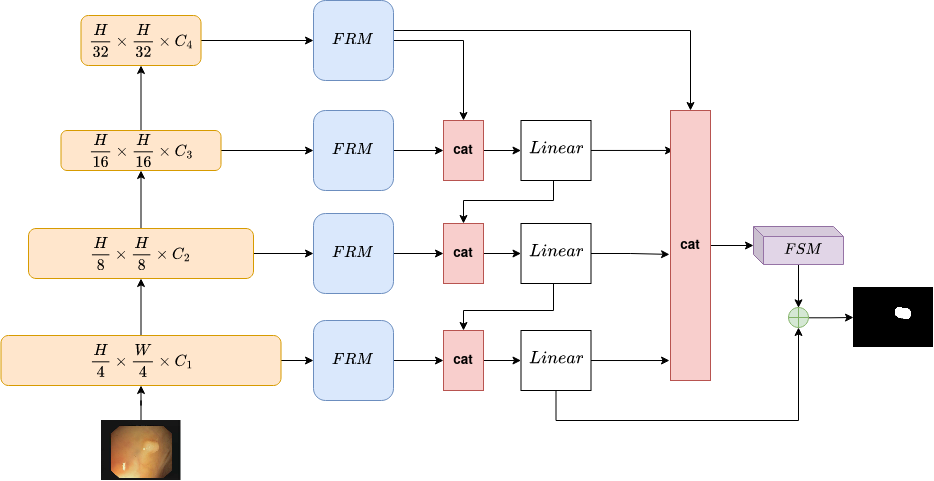}
    \caption{The architecture of proposed LAPFormer. Image is fed into the Encoder, which is MiT. Features from all stages are enhanced with Feature Refinement Module in the Decoder. Progressive Feature Fusion is performed to slowly fuse features from different scales, then all are aggregated and combine with low level features from the feature hierarchy through a skip-connection to perform prediction. "FRM" is Feature Refinement Module, "cat" denotes the concatenate operation, "Linear" denotes the MLP, "FSM" is the Feature Selection Module and "+" denotes the addition operation.}
    \label{fig:Model}
\end{figure*}

In this section, we describe the proposed LAPFormer in detail. An overview of our model is presented in Fig \ref{fig:Model}.
\subsection{Encoder}
We choose MiT (Mix Transformer) proposed in \cite{xie2021segformer} as our segmentation encoder for two main reasons. First, it consists of a novel hierarchically structured Transformer encoder which produces multi-level multi-scale feature outputs. Second, MiT uses a convolutional kernel instead of Positional Encoding (PE), therefore avoiding decreased performance when the testing resolution differs from training. Convolutional layers are argued to be more adequate for extracting location information for Transformer than other methods. Furthermore, MiT uses small image patches of size $4 \times 4$, which are proved to favor intensive dense prediction tasks like semantic segmentation. 

Assume that we have input image $X$ with spatial dimensions of $H \times W \times 3$ (represent the height, width and channel). MiT generates four different level feature $f_i$ with resolution $\frac{H}{2^{i+1}} \times \frac{W}{2^{i+1}} \times C_i$ where $i\in \{{1, 2, 3, 4}\}$ and $C_{i+1}$ is larger than $C_i$. Basically, MiT has six variants with same architecture but varies in sizes, from MiT-B0 to MiT-B5.

\subsection{Progressive Feature Fusion}

Dense prediction is a famous task in computer vision which consists of Object Detection, Semantic Segmentation, ... Object of vastly different scales all appear in the same picture. Therefore,  multi-scale features are heavily required to generate good results. The most popular way to utilize multi-scale features is to construct a feature pyramid network (FPN) \cite{lin2017feature}. However, as pointed out in \cite{wang2020scale} and \cite{pang2019libra}, there are big semantic gaps between feature from non-adjacent scales, features from two scale levels apart are weakly correlated, while features between adjacent scales are highly correlated.  

In FPN, features from upper scale are upscaled then directly added to features from lower scale. We argue that this is sub-optimal, when fusing features, they should be further processed before directly fusing to next level. We propose Progressive Feature Fusion module, which progressively fuses features from upper scales to lower scales, therefore, reduce the information gap between the low resolution, high semantic feature maps and high resolution, low semantic ones.  

Instead of fusing feature maps with addition operation, we use concatenation operation. Feature maps of all scale are upsampled to $(H/4, W/4)$. Then feature maps from upper scale are progressively fused with lower scale as follow:

$$
F(x_{i}, x_{i-1}) = Linear([x_i, x_{i-1}])
$$

where $x_i$ is features from scale $i$; $x_{i-1}$ is features from one scale lower than $i$; $[ \dots ]$ is concatenation operation; \textit{Linear} is a fully-connected layer.  

\subsection{Aggregation for Prediction}

\begin{figure}
\centering
\begin{subfigure}{.24\textwidth}
  \centering
  \includegraphics[height=4cm, width=2.7cm]{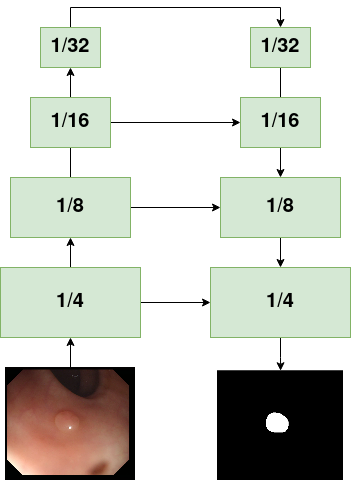}
  \caption{Conventional Encoder-Decoder architecture}
  \label{ConventionalDecoder}
\end{subfigure}
\hfill
\begin{subfigure}{.24\textwidth}
  \centering
  \includegraphics[height=4cm, width=3.5cm]{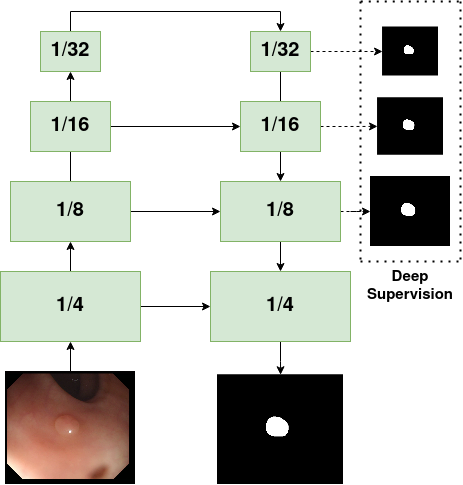}
  \caption{Decoder with Deep Supervision architecture}
  \label{Deepsupervision}
\end{subfigure}
\hfill
\begin{subfigure}{.24\textwidth}
  \centering
  \includegraphics[height=4cm, width=4cm]{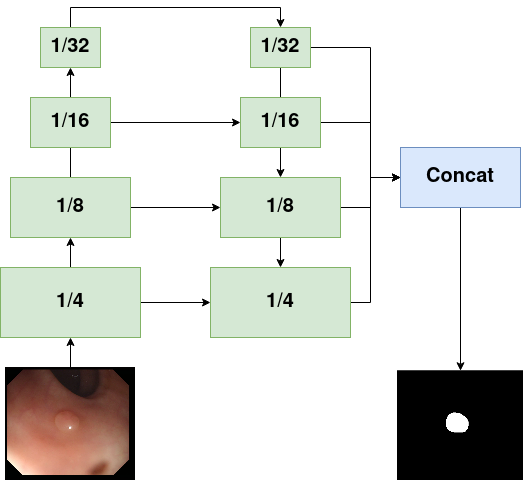}
  \caption{Our proposed Aggregation for prediction}
  \label{ProposedDecoder}
\end{subfigure}
\caption{A comparison between conventional other semantic segmentation architecture and our proposed architecture, (a) Encoder-Decoder model, (b) Decoder with Deep Supervision, (c) Aggregation for Prediction (ours)}
\label{fig:Decoders}
\end{figure}

On conventional Encoder-Decoder architecture, after performing multi-scale feature fusion from feature pyramid, other models \cite{ronneberger2015u}, \cite{oktay2018attention}, \cite{jha2019resunet++}, \cite{wang2022stepwise} often perform prediction on the last output feature maps, which has the highest resolution (Fig \ref{ConventionalDecoder}). Others adopt auxiliary predictions during training \cite{kim2021uacanet}, \cite{lou2021caranet} but during test time, they only make prediction from the highest resolution feature map (Fig \ref{Deepsupervision}). 

We argue that predicting on only the highest resolution feature maps is sub-optimal. This make the highest resolution feature maps to carry a lot of information, but information can be lost during feature fusion process. Adding information from high semantic branches (Figure \ref{ProposedDecoder}) before prediction can provide network with coarse polyp regions, therefore, easing the work of highest resolution feature maps. 

\subsection{Feature Selection Module}

\begin{figure}
    \centering
    \includegraphics[width=0.9\linewidth]{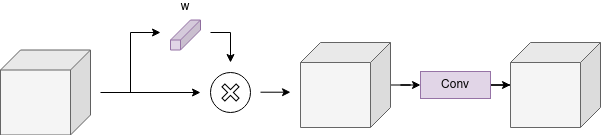}
    \caption{Feature Selection Module}
    \label{fig:FSM}
\end{figure}

After channel concatenation to increase information for prediction, it is crucial to emphasize feature maps that contain important details and suppress redundant information.  

The data flow of Feature Selection Module is shown on Fig \ref{fig:FSM}. Output feature maps from Aggregation for Prediction is scaled with a weighting vector $w$ which indicates the importance of each channel. Then we conduct channel reduction to keep the importance ones for prediction. The weighting vector $w$ is calculated as:

$$
w = \sigma(fc_2(\delta(fc_1(x))))
$$

where $x$ is obtained by applying average pooling on output feature maps from Aggregation for Prediction mentioned above; $fc_1, fc_2$ are two fully-connected layers; $\delta$ is ReLU activation function and $\sigma$ is Hard Sigmoid activation function.  

Overall, the process of Feature Selection Module (FSM) can be formulated as:

$$
X_{inter} = w \cdot X
$$
$$
X_{pre} = Conv(X_{inter})
$$

where $X$ is output feature maps from Aggregation for Prediction; $w$ is weighting vector; $X_{inter}$ is intermediate feature maps; $X_{pre}$ is output of FSM module.

\subsection{Low level connection}

Low level features are important in semantic segmentation task, and its importance is especially high in polyp segmentation, because the boundary between polyp and its surroundings are not easily distinguishable.  

We further enhance prediction on polyp boundary using a skip-connection. The skip-connection connects output feature maps from FSM with the lowest level feature maps in the feature hierarchy from Progressive Feature Fusion. This is a simple but surprisingly effective way which boosts the performance of our model. 

\subsection{Feature Refinement Module}

Normally, feature maps of every scale from backbone pass through a lateral connection. It merges feature maps from upper scale into lower scale feature maps. Lateral connection is implemented using a $1 \times 1$ Convolution. In Object Detection settings, this isn't a problem, because features after feature fusion continue to be refined in the Detection Head.  

However, in our Decoder settings, this is sub-optimal for multiple reasons: 

\begin{itemize}
    \item All module in our Decoder part use $1 \times 1$ Convolution. This hampers the learnability of model in the Decoder.
    \item $1 \times 1$ Convolution in lateral connection reduces number of channels in two top-most scale. This process can cause information lost.  
    \item $1 \times 1$ Convolution in lateral connection also increases number of channels in two bottom-most scale. This process can cause information redundancy. 
\end{itemize}

\begin{figure}
    \centering
    \includegraphics[width=0.9\linewidth]{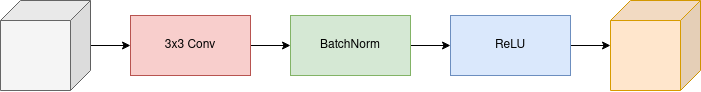}
    \caption{Feature Refinement Module}
    \label{fig:FRM}
\end{figure}

We propose to replace lateral connection with our Feature Refinement Module (FRM). Backbone feature maps from different levels are refined with FRM. It should be noted that each level has its own respective FRM. FRM is implemented using a  $3 \times 3$ Convolution, followed by a BatchNorm layer and ReLU activation function (Fig \ref{fig:FRM}). This is very simple but harmonize well with our Encoder and Decoder architecture. Using $3 \times 3$ Convolution brings two benefits:

\begin{itemize}
    \item $3 \times 3$ Convolution further refines features coming from backbone, therefore, enhances the learning capability of Decoder.  
    \item Transformer Encoder lacks Local Receptive Field, result in lacking of local details. $3 \times 3$ Convolution effectively emphasizes local features and cleans up noises.
\end{itemize}

\section{Experiments}

\textbf{Dataset and Evaluation Metrics:} We construct experiments on five polyp segmentation datasets: Kvasir \cite{jha2020kvasir}, CVC-ClinicDB \cite{bernal2015wm}, CVC-ColonDB \cite{tajbakhsh2015automated}, CVC-T \cite{vazquez2017benchmark} and ETIS-Larib Polyp DB \cite{silva2014toward}. We follow the experimental scheme mentioned in PraNet \cite{fan2020pranet} and UACANet \cite{kim2021uacanet} which randomly extract 1450 images both from Kvasir and CVC-ClinicDB to construct a training dataset. We used the same training dataset as in PraNet and UACANet. Then we perform evaluation on the rest of Kvasir and CVC-ClinicDB. We also evaluate on CVC-ColonDB, CVC-T and ETIS to show our model's generalization ability on unseen datasets. For performance measuring, we use mean Dice and mean IoU score as evaluation metrics for our experiments.

\textbf{Implementation details:} Our implementation is based on PyTorch and MMSegmentation \cite{mmseg2020} toolbox. Training is performed on Google Colab virtual machine, with a NVIDIA Tesla V100 GPU and 16GB RAM. We used AdamW optimizer with initial learning rate of 0.0001, along with Cosine Annealing \cite{loshchilov2016sgdr} scheduler. We resize images to $352^2$ for training and testing. For data augmentations, we employ flip, slight color jittering and cutout \cite{devries2017improved}. Our loss function is a combination of Binary Cross Entropy and Dice Loss. Our model is trained 5 times for 50 epochs with batch size of 16. Reported results are averaged over 5 runs. 

\subsection{Ablation Study}

For ablation study, we use MiT-B1 backbone and train model for 50 epochs average over 5 runs. All results are reported under Table \ref{Ablation}

\begin{table*}[t]
\centering
\caption{Ablation study on each component}
\begin{tabular}{|l|l|l|llllllllll|}
\hline

\multirow{2}{*}{Methods} & \multirow{2}{*}{GFLOPS} & \multirow{2}{*}{Params (M)} & \multicolumn{2}{c|}{Kvasir} & \multicolumn{2}{c|}{ClinicDB} & \multicolumn{2}{c|}{ColonDB} & \multicolumn{2}{c|}{CVC-T} & \multicolumn{2}{c|}{ETIS} \\ \cline{4-13} 

 &  &  & \multicolumn{1}{c}{mDice} & \multicolumn{1}{c|}{mIoU} & \multicolumn{1}{c}{mDice} & \multicolumn{1}{c|}{mIoU} & \multicolumn{1}{c}{mDice} & \multicolumn{1}{c|}{mIoU} & \multicolumn{1}{c}{mDice} & \multicolumn{1}{c|}{mIoU} & \multicolumn{1}{c}{mDice} & \multicolumn{1}{c|}{mIoU} \\ \hline

SegFormer Head & 5.5 & 13.68 & 0.902 & \multicolumn{1}{l|}{0.844} & \textbf{0.904} & \multicolumn{1}{l|}{\textbf{0.851}} & 0.754 & \multicolumn{1}{l|}{0.667} & 0.838 & \multicolumn{1}{l|}{0.764} & 0.753 & 0.672 \\
PFF & 6.53 & 13.81 & 0.904 & \multicolumn{1}{l|}{0.846} & 0.895 & \multicolumn{1}{l|}{0.838} & 0.765 & \multicolumn{1}{l|}{0.676} & 0.846 & \multicolumn{1}{l|}{0.769} & 0.753 & 0.669 \\
PFF + A & 8.57 & 14.07 & 0.903 & \multicolumn{1}{l|}{0.844} & 0.895 & \multicolumn{1}{l|}{0.840} & 0.769 & \multicolumn{1}{l|}{0.680} & 0.852 & \multicolumn{1}{l|}{0.774} & 0.759 & 0.676 \\
PFF + A + FSM & 8.58 & 14.21 & 0.906 & \multicolumn{1}{l|}{0.851} & 0.897 & \multicolumn{1}{l|}{0.841} & 0.769 & \multicolumn{1}{l|}{0.680} & 0.855 & \multicolumn{1}{l|}{0.776} & 0.759 & 0.675 \\
PFF + A + FSM + Skip & 8.58 & 14.21 & 0.904 & \multicolumn{1}{l|}{0.846} & 0.898 & \multicolumn{1}{l|}{0.844} & 0.774 & \multicolumn{1}{l|}{0.687} & 0.858 & \multicolumn{1}{l|}{\textbf{0.781}} & 0.764 & 0.682 \\
LAPFormer Head & 10.54 & 16.3 & \textbf{0.910} & \multicolumn{1}{l|}{\textbf{0.857}} & 0.901 & \multicolumn{1}{l|}{0.849} & \textbf{0.780} & \multicolumn{1}{l|}{\textbf{0.695}} & \textbf{0.859} & \multicolumn{1}{l|}{0.781} & \textbf{0.768} & \textbf{0.686} \\ \hline
                                               
\end{tabular}
\label{Ablation}
\end{table*}

\begin{figure}{}
    \centering
    \includegraphics[width=0.49\textwidth]{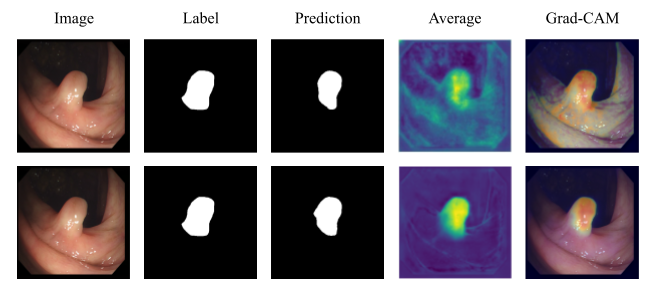}
    \caption{Influence of Progressive Feature Fusion (PFF). Top row: base SegFormer Head without PFF. Second row: Proposed Progressive Feature Fusion. "Average" is average pooling across locations of the feature maps. "Grad-CAM" is class activation mapping obtained using Grad-CAM}
    \label{fig:Ablation_PFF}
\end{figure}

\textbf{Progressive Feature Fusion (PFF) and Aggregation for Prediction.} We demonstrate the effect of Progressive Feature Fusion (PFF) in Fig \ref{fig:Ablation_PFF}. Visualization is done on feature maps closest to prediction head. After applying PFF, we can see that our model accurately localizes Polyp region and clears up noises from other regions. PFF effectively boosts performance of our Encoder's across almost all datasets. Aggregating all feature maps produced by PFF helps utilize multi-scale features better and brings huge performance gain on ETIS, which consists of many small polyps. Results for both of these components are shown on the second and third row of Table \ref{Ablation}

\begin{figure}
    \centering
    \includegraphics[width=0.49\textwidth]{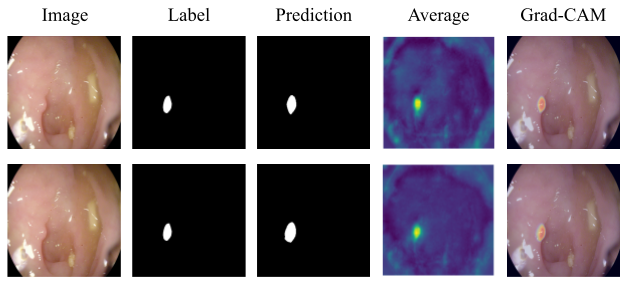}
    \caption{Effectiveness of Feature Selection Module (FSM). Top row: Our PFF + A components in Decoder. Second row: PFF + A with addition FSM in Decoder. "Average" is average pooling across locations of the feature maps. "Grad-CAM" is class activation mapping obtained using Grad-CAM}
    \label{fig:Ablation_FSM}
\end{figure}

\textbf{Feature Selection Module (FSM).} Applying FSM before giving prediction helps the prediction head to focus on important regions. The strength of FSM is demonstrated on Fig \ref{fig:Ablation_FSM}. With the same output feature maps from PFF + A, adding FSM refines the polyp segmentation region, highlights details which were not captured before. The result of FSM is shown on the forth row of Table \ref{Ablation} 

\begin{figure}
    \centering
    \includegraphics[width=0.49\textwidth]{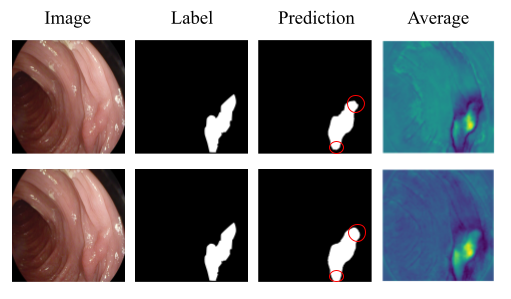}
    \caption{Impact of low level features on boundary regions. Top row: PFF + A + SFM in Decoder. Second row: PFF + A + SFM and an addition skip connection from low level feature maps. "Average" is average pooling across locations of the feature maps.}
    \label{fig:skip}
\end{figure}

\textbf{Low level connection.} Low level features have always been important in dense prediction tasks. For polyp segmentation, integrating low level features produces finer segmentation masks, especially the boundary regions. The importance of low level features is shown on Fig \ref{fig:skip}, adding more information from low level features refines polyp segmentation mask. This low level connection consistently boosts our model performance on almost all datasets, results are shown on the fifth row of Table \ref{Ablation}. 

\textbf{Feature Refinement Module (FSM).} Incoperating Feature Refinement Module (FSM) increases our model's performance on all datasets. Results after applying FSM is shown on last row of Table \ref{Ablation}. With FSM, we are able to scale our model with stronger backbone. All variations of our model are shown on Table \ref{results_all}.

\begin{table*}[]
\centering
\caption{Evaluation on different variations of LAPFormer}
\begin{tabular}{|l|ll|ll|ll|ll|ll|}
\hline
\multirow{2}{*}{Methods} & \multicolumn{2}{c|}{Kvasir} & \multicolumn{2}{c|}{ClinicDB} & \multicolumn{2}{c|}{ColonDB} & \multicolumn{2}{c|}{CVC-T} & \multicolumn{2}{c|}{ETIS} \\ \cline{2-11} 
 & \multicolumn{1}{c}{mDice} & \multicolumn{1}{c|}{mIoU} & \multicolumn{1}{c}{mDice} & \multicolumn{1}{c|}{mIoU} & \multicolumn{1}{c}{mDice} & \multicolumn{1}{c|}{mIoU} & \multicolumn{1}{c}{mDice} & \multicolumn{1}{c|}{mIoU} & \multicolumn{1}{c}{mDice} & \multicolumn{1}{c|}{mIoU} \\ \hline
LAPFormer-S & 0.910 & 0.857 & 0.901 & 0.849 & 0.781 & 0.695 & 0.859 & 0.781 & 0.768 & 0.686 \\
LAPFormer-M & 0.913 & 0.863 & 0.911 & 0.861 & 0.800 & 0.722 & 0.856 & 0.790 & 0.784 & 0.709 \\
LAPFormer-L & \textbf{0.917} & \textbf{0.866} & \textbf{0.915} & \textbf{0.866} & \textbf{0.815} & \textbf{0.735} & \textbf{0.891} & \textbf{0.821} & \textbf{0.797} & \textbf{0.720} \\ \hline
\end{tabular}
\label{results_all}
\end{table*}

\subsection{Comparison with State-of-the-Art}

We compare our results with existing approaches on 5 benchmark datasets. Table \ref{compare} shows the results of SOTA methods. Our model achieves competitive performance on all datasets while being incredibly lighter than other methods. FLOPs and Params are shown on Table \ref{FLOPs}.

\begin{table*}[t]
\centering
\caption{Comparison with other approaches on 5 benchmark datasets}
\begin{tabular}{|c|cc|cc|cc|cc|cc|}
\hline
\multirow{2}{*}{Method} & \multicolumn{2}{c|}{Kvasir} & \multicolumn{2}{c|}{ClinicDB} & \multicolumn{2}{c|}{ColonDB} & \multicolumn{2}{c|}{CVC-T} & \multicolumn{2}{c|}{ETIS} \\ \cline{2-11} 
                        & mDice        & mIoU         & mDice         & mIoU          & mDice         & mIoU         & mDice        & mIoU        & mDice       & mIoU        \\ \hline
PraNet \cite{fan2020pranet}                  & 0.898        & 0.840        & 0.899         & 0.849         & 0.709         & 0.640        & 0.871        & 0.797       & 0.628       & 0.567       \\
Polyp-PVT \cite{dong2021polyp}               & 0.917        & 0.864        & 0.937         & 0.889         & 0.808         & 0.727        & 0.900        & 0.833       & 0.787       & 0.706       \\
SANet \cite{wei2021shallow}                   & 0.904        & 0.847        & 0.916         & 0.859         & 0.753         & 0.670        & 0.888        & 0.815       & 0.750       & 0.654       \\
MSNet \cite{zhao2021automatic}                  & 0.907        & 0.862        & 0.921         & 0.879         & 0.755         & 0.678        & 0.869        & 0.807       & 0.719       & 0.664       \\
TransFuse-L* \cite{zhang2021transfuse}            & 0.920        & 0.870        & \textbf{0.942}         & \textbf{0.897}         & 0.781         & 0.706        & 0.894        & 0.826       & 0.737       & 0.663       \\
SSFormer-L \cite{wang2022stepwise}             & 0.917        & 0.864        & 0.906         & 0.855         & 0.802         & 0.721        & 0.895        & 0.827       & 0.796       & 0.720       \\
ColonFormer-L \cite{duc2022colonformer}          & \textbf{0.924}        & \textbf{0.876}        & 0.932         & 0.884         & 0.811         & 0.733        & \textbf{0.906}        & \textbf{0.842}       & \textbf{0.801}       & \textbf{0.722}       \\
ColonFormer-XL \cite{duc2022colonformer}         & 0.920        & 0.870        & 0.923         & 0.875         & 0.814         & \textbf{0.735}        & 0.905        & 0.840       & 0.795       & 0.715       \\
LAPFormer-S               & 0.910        & 0.857        & 0.901         & 0.849         & 0.781         & 0.695        & 0.859        & 0.781       & 0.768       & 0.686       \\
LAPFormer-M               & 0.913        & 0.863        & 0.911         & 0.861         & 0.800         & 0.722        & 0.856        & 0.790       & 0.784       & 0.709       \\
LAPFormer-L               & 0.917        & 0.867        & 0.915         & 0.866         & \textbf{0.815}         & \textbf{0.735}        & 0.891        & 0.821       & 0.797       & 0.719       \\ \hline
\end{tabular}
\label{compare}
\end{table*}

\begin{table}[]
\centering
\caption{Number of Parameters and FLOPs of different methods}
\begin{tabular}{|l|l|l|}
\hline
\multirow{2}{*}{Methods} & \multirow{2}{*}{GFLOPs} & \multirow{2}{*}{Params (M)} \\
 &  &  \\ \hline
PraNet \cite{fan2020pranet} & 13.11 & 32.55 \\
CaraNet \cite{lou2021caranet} & 21.69 & 46.64 \\
TransUNet \cite{chen2021transunet} & 60.75 & 105.5 \\
ColonFormer-S \cite{duc2022colonformer} & 16.03 & 33.04 \\
ColonFormer-L \cite{duc2022colonformer} & 22.94 & 52.94 \\
SSFormer-S \cite{wang2022stepwise} & 17.54 & 29.31 \\
SSFormer-L \cite{wang2022stepwise} & 28.26 & 65.96 \\
\textbf{LAPFormer-S} & \textbf{10.54} & \textbf{16.3} \\
\textbf{LAPFormer-L} & \textbf{18.96} & \textbf{47.22} \\ \hline
\end{tabular}
\label{FLOPs}
\end{table}

\section{Conclusion}
In this work, we propose a novel light and accurate Transformer-based model for Polyp segmentation called LAPFormer. The decoder part of LAPFormer is capable of capturing not only rich semantic features for coarse polyp region but also boundary regions of polyp to produce a fine segmentation mask. Extensive experiments show that our model achieves competitive results while being significantly less in computation cost. 

\section{Acknowledge}
This work is partially supported by \textbf{\textit{Sun-Asterisk Inc}}. We would like to thank our colleagues at \textbf{\textit{Sun-Asterisk Inc}} for their advice and expertise. Without their support, this experiment would not have been accomplished.

\bibliographystyle{IEEEtran}
\bibliography{references}

\end{document}